\pdfoutput=1
\documentclass[11pt]{article}

\usepackage[preprint]{acl}

\usepackage{booktabs}

\usepackage{times}
\usepackage{latexsym}
\usepackage{amsmath}

\usepackage[T1]{fontenc}
\usepackage[utf8]{inputenc}

\usepackage{microtype}

\usepackage{inconsolata}

\usepackage{graphicx}
\usepackage{cleveref}

\usepackage{adjustbox}

\usepackage{siunitx}
\usepackage{multicol}
\usepackage{multirow}
\usepackage{array}
\usepackage{enumitem}

\definecolor{darkgreen}{rgb}{0.0, 0.5, 0.0}

\newcommand\sect[1]{\S\ref{#1}}

\title{Characterizing Linguistic Shifts in Croatian News via \\ Diachronic Word Embeddings}
\author{David Dukić \and Ana Barić \and Marko Čuljak \and Josip Jukić \and Martin  Tutek\\ TakeLab, Faculty of Electrical Engineering and Computing, University of Zagreb \\ \texttt{\{name.surname\}@fer.hr} \\}

\begin{document}
\maketitle

\begin{abstract}

Measuring how semantics of words change over time improves our understanding of how cultures and perspectives change.
Diachronic word embeddings help us quantify this shift, although previous studies leveraged substantial temporally annotated corpora. %, which are rare.
In this work, we use a corpus of $9.5$ million Croatian news articles spanning the past $25$ years and quantify semantic change using skip-gram word embeddings trained on five-year periods.
Our analysis finds that word embeddings capture linguistic shifts of terms pertaining to major topics in this timespan (\textsc{covid-19}, Croatia joining the European Union, technological advancements).
We also find evidence that embeddings from post-2020 encode increased positivity in sentiment analysis tasks, contrasting studies reporting a decline in mental health over the same period.\footnote{
\url{https://github.com/dd1497/cro-diachronic-emb}} % effects from media narratives could ripple into unexpected domains. \david{align with sentiment shift findings}
\end{abstract}

\section{Introduction}

% The short papers may be anonymous, but that is not required.
% [Short papers] should not exceed four (4) pages for content plus two (2) pages for references.

The progress of culture and technology is reflected in language, which adapts to incorporate novel meanings into existing words or by entirely changing their semantics.
Such changes exhibit systematic regularities with respect to word frequency and polysemy \citep{breal1904essai,ullman1962introduction}, and can be detected by studies on distributed word representations \citep{hamilton-etal-2016-diachronic}.
Studies of diachronic word embeddings have detected known changes in word meaning in English-language books spanning multiple centuries. However, such analyses are limited to languages historically abundant in text corpora, as learning high-quality distributed word representations requires diverse contexts.
In our work, we rely on a Croatian online news corpus containing articles from the last $25$ years \citep{dukic2024takelab}. We investigate whether major topics in this period are reflected in word semantics and evaluate the practical implications of semantic shift on the use case of sentiment analysis.

\begin{figure}
  \centering
  \includegraphics[width=\columnwidth]{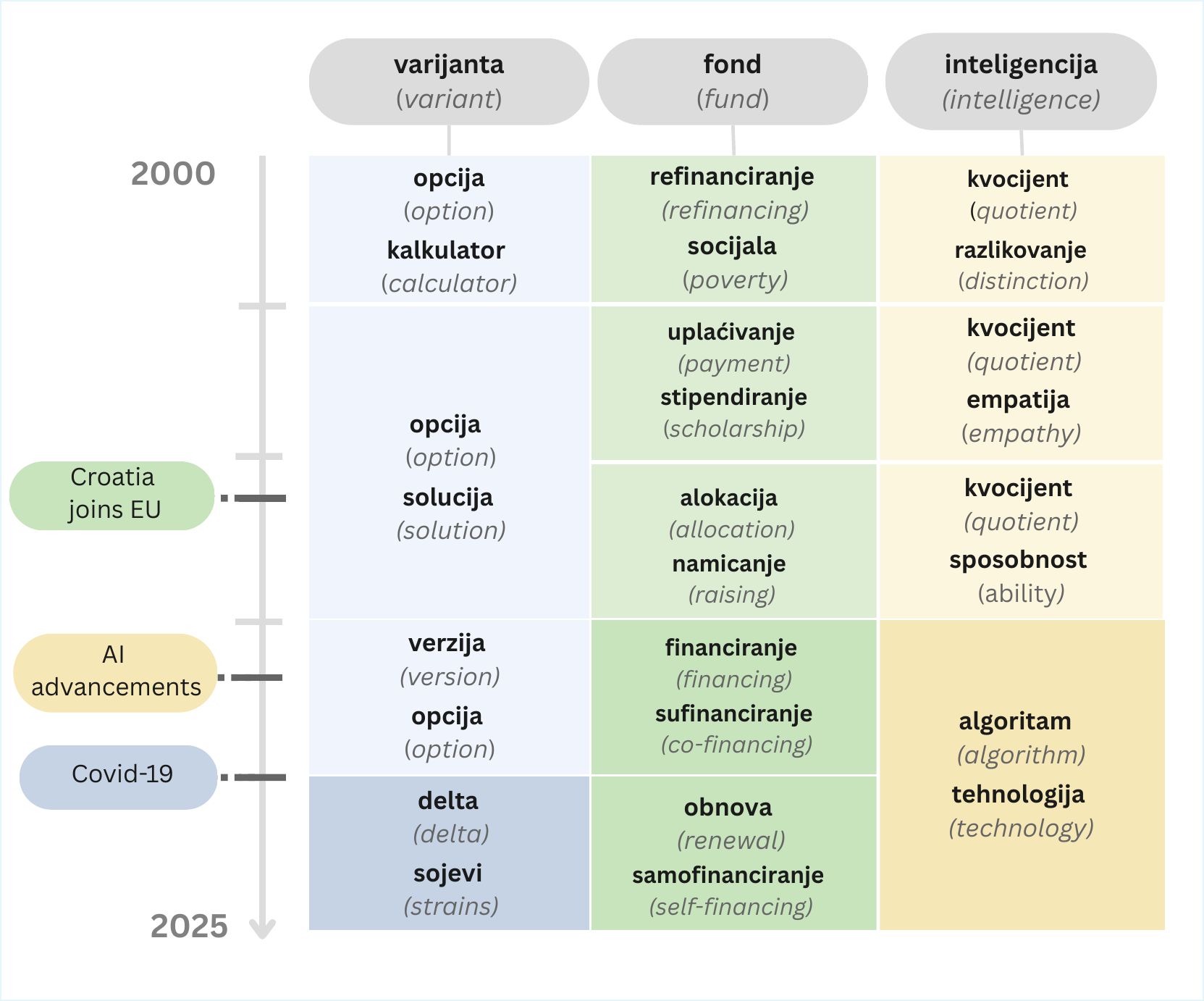} % Auto-fits to column width
  %\vspace{-2em}
  \caption{Linguistic shifts in Croatian news outlets over $25$ years, driven by three major events: EU membership in 2013, technological progress in 2017, and COVID-19 in 2020.}
  %\vspace{-1.5em}
  \label{fig:column-fit}
\end{figure}

We split the corpus into five periods of equal duration, train distributed word representations \citep{mikolov2013distributed} for each period, and verify their quality. Next, we select three major topics that likely influenced the meaning of Croatian words during these periods and semi-automatically curate a list of related words for each topic. We show that these words undergo strong linguistic shifts \citep{hamilton-etal-2016-cultural}, acquiring new meanings and demonstrating the rapid impact of narrative on distributional semantics (see \Cref{fig:column-fit}).

To evaluate whether linguistic shifts affect word representations in practice, we first align word embeddings from different periods, then transfer such aligned embeddings onto a model based on embeddings from another period and observe the change in average predicted sentiment intensity.
We find that embeddings from later periods are \textit{more positive} despite studies showing that mental health has been negatively affected \citep{rozanov2019mental,cullen2020mental}.
In short, our contributions are as follows: (1) We train diachronic word embeddings on a corpus of Croatian news articles, which we make available for further studies;\footnote{\url{https://www.takelab.fer.hr/retriever/cro-diachronic-emb.zip}} (2) We show that corpora spanning short timespans accurately reflect major topics through linguistic shifts of associated words; (3) We find that the sentiment of word embeddings trained on news corpora becomes more positive in recent periods.

\section{Related Work}
Various studies explore word embeddings as a diachronic tool \citep[][\textit{inter alia}]{hamilton-etal-2016-cultural, hamilton-etal-2016-diachronic, schlechtweg-etal-2019-wind, fivser2019distributional, kurtygit2021lexical, schlechtweg2024sense}.
By leveraging methods from distributional semantics, which encode individual words in vector spaces based on co-occurrence \citep{mikolov2013distributed}, researchers study how global and local neighborhoods of individual words change over time \citep{hamilton-etal-2016-diachronic}.
There is a variety of causes driving semantic shift, with two major ones being \textit{linguistic shift}, where words take on a new meaning while retaining previous ones, and \textit{cultural shift}, where technological progress completely alters the way a word is used \citep{hamilton-etal-2016-cultural}. In our work, we follow the methodology used by \citet{hamilton-etal-2016-diachronic}, apply it to a corpus of Croatian newswire texts, and extend the setup to evaluate practical effects of \textit{linguistic shift} on major topics and sentiment analysis.

The majority of diachronic embedding studies explore corpora spanning several centuries, grounded in books \citep{hamilton-etal-2016-diachronic, hamilton-etal-2016-cultural, schlechtweg-etal-2019-wind, kurtygit2021lexical}.
Due to the lack of such corpora of sufficient scale in Croatian,
% As such corpora in Croatian lack sufficient scale for meaningful analyses based on distributional semantics, 
we leverage a recently introduced dataset of Croatian newswire corpora \citep{dukic2024takelab}, which covers a shorter period of $25$ years. % as a trade-off for scale.
Despite the narrower timeframe, we hypothesize that the corpus sufficiently captures the diachronic shift in word meaning, which we experimentally verify in this work.

\section{Methodology}
% In this section, we first delineate our setup for learning diachronic word embeddings (\sect{sec:diachronic}), estimating their quality (\sect{sec:quality}), and then explain the analyses we conduct on topical linguistic shift (\sect{sec:topical-shift}) and sentiment shift (\sect{sec:senti-shift}). % \david{this is a candidate for slacking}

\subsection{Diachronic Word Embeddings}
\label{sec:diachronic}

\paragraph{Dataset.} We train word embeddings on the TakeLab Retriever corpus of Croatian newswire articles \citep{dukic2024takelab}. The corpus consists of \num{9450929} articles crawled from $33$ Croatian news outlets across $25$ years (2000--2024) and contains around $3.7$ billion words (see Table \ref{tab:data_stats} for more details).
% The initial corpus also contained the \textit{low-quality articles} (see \citet{dukic2024takelab}), which we discarded.
We use spaCy \verb|hr_core_news_lg| \citep{spacy} to sentencize, tokenize, and tag parts of speech in the corpus.
% For word processing, we rely on the spaCy model \verb|hr_core_news_lg| \citep{spacy} to split articles into sentences and sentences into words.
As the Croatian language is highly inflectional, we lemmatize the corpus with the lexicon-based MOLEX lemmatizer \citep{vsnajder2008automatic} and differentiate between homonyms with part-of-speech tags obtained using the Croatian spaCy tagger applied to raw words from articles. We split the corpus into $5$ five-year periods.

\paragraph{Method.} We use the skip-gram with negative sampling (SGNS) method from Word2Vec \citep{mikolov2013distributed} to train our word embeddings. We use the \textsc{gensim} implementation of SGNS to train the embeddings \citep{rehurek_lrec}. We list the hyperparameter values and hardware details in \autoref{app:hyperparams}.
\begin{table}
\centering
\small{\begin{tabular}{crr}
\toprule
\textbf{Period} & \textbf{\#Words} & \textbf{\#Unique words} \\
\midrule
1 (2000--2004) & \num{53062322} & \num{589769} \\
2 (2005--2009) & \num{158028732} & \num{1191784} \\
3 (2010--2014) & \num{551701502} & \num{2583363} \\
4 (2015--2019) & \num{1170882497} & \num{3462601} \\
5 (2020--2024) & \num{1753495356} & \num{3975631} \\
\midrule
\textbf{Total} & \num{3687170409} & \num{11803148} \\
\bottomrule
\end{tabular}}
\caption{The number of words and unique words per $5$ five-year periods in the Croatian online news corpus.}
\label{tab:data_stats}
% \vspace{-1.5em}
\end{table}

\subsection{Embedding Quality}
\label{sec:quality}
We validate the quality of the learned embeddings on two word similarity corpora for Croatian: CroSemRel450 \citep{jankovic2011random} and CroSYN \citep{snajder2013building}. CroSemRel450 contains human-annotated pairs of words rated for semantic relatedness, while CroSYN is a synonym choice dataset comprising one correct synonym and three unrelated options for each target word. %VerbCROcean \citep{sekulic-snajder-2016-verbcrocean} provides fine-grained semantic relations among Croatian verbs.

\subsection{Topical Linguistic Shift}
\label{sec:topical-shift}

We hypothesize that diachronic embeddings over periods can reveal significant topical linguistic shifts.
To unveil these shifts, we curate words pertaining to three major topics relevant globally and/or to Croatia: the \textsc{covid-19} crisis, Croatia joining the \textit{European Union (EU)}, and \textit{technological progress}.
We expect \textsc{covid-19} to produce the highest shift in the fifth period, joining the EU in the second, third, and fourth periods (as Croatia entered the EU in 2013), and technological progress in the fourth and fifth periods (digitalization after entering EU and proliferation of AI in the fifth period).
Finding no substantial shifts for verbs or adjectives, we focus on the change in nouns as they are more prone to linguistic shifts \citep{hamilton-etal-2016-cultural}. We measure the shift of each word using the cumulative shift score, based on the halved cosine distance ($\cos$) over neighboring periods: 
\[
     D_\mathrm{c} = \sum_{i=1}^4 \frac{1 - \cos(\mathbf{v}_{i}, \mathbf{v}_{i+1})}{2}.
\]
For this analysis, we use Procrustes alignment \citep{schonemann1966generalized} to align word embeddings across periods. We begin by recursively aligning pairs of embeddings, starting from the most recent, fifth period (2020-2024), and then moving toward the earlier ones.
%We begin by aligning the embeddings from the fifth period to those of the fourth. We then proceed to align the obtained embeddings with the ones from the third period, repeating this process for periods two and one.
Let $\mathbf{E}_t$ denote the embedding matrix for period $t$, and let $\mathrm{PA}(\mathbf{A}, \mathbf{B})$ denote the Procrustes alignment of matrix $\mathbf{A}$ to $\mathbf{B}$.
We use $\mathbf{E}_t^{\ast}$ to denote the aligned embeddings for period $t$.
The alignment procedure can be written recursively as:
\[
\begin{aligned}
\mathbf{E}_t^* &=
\begin{cases}
\mathbf{E}_5, & \text{if } t = 5,\\
\mathrm{PA}\bigl(\mathbf{E}_{t+1}^*,\,\mathbf{E}_{t}\bigr), & \text{if } t \in \{1,2,3,4\}.
\end{cases}
\end{aligned}
\]
%Denoting $\mathrm{PA}$ as Procrustes alignment and $\mathbf{E}_x$ as embedding matrix of the shared vocabulary between periods for period $x$, the aligned embeddings are obtained as $\mathrm{PA}(\mathrm{PA}(\mathrm{PA}(\mathrm{PA}(\mathbf{E}_5,\mathbf{E}_4), \mathbf{E}_3), \mathbf{E}_2), \mathbf{E}_1)$.
Further details are provided in \Cref{subsec:tls_res}.
%\footnote{\footnotesize{We also explore other word types, such as verbs and adjectives, but do not find meaningful shifts.}}

% \noindent where the cosine similarity between two word embeddings $\mathbf{v}_i$ and $\mathbf{v}_j$ is defined as $\cos(\mathbf{v}_i, \mathbf{v}_j) = \frac{\mathbf{v_i} \cdot \mathbf{v_j}}{\|\mathbf{v_i}\| \|\mathbf{v_j}\|}$.

\subsection{Sentiment Shift}
\label{sec:senti-shift}
Distributed word representations capture contextual cues helpful in determining the tone and sentiment of texts, serving as a more robust and effective alternative to lexicon-based and traditional machine learning approaches \cite{zhang2018deep,alsaqqa2020use,wankhade2022survey}.
To quantify sentiment shifts in our corpus, we train a classifier $C_i$ for each period $t_i$  using embeddings $E_i$ computed on the corpus from $t_i$. Each classifier predicts the sentiment label (positive, neutral, or negative) of a text sequence based on the average of the word embeddings within the sequence. Next, we compute the average sentiment of a classifier $C_i$ on a test set using the word embeddings from $E_i$ and denote this quantity by $\bar{s}_{i\leftarrow i}$. We repeat the same procedure for $C_i$ with Procrustes-aligned embeddings from each other period $E_j^*, \; j \neq i$ to obtain quantities $\bar{s}_{i\leftarrow j}$.
% for every period $t_j, j \neq i$ and
% Since $E_i$ and $E_j$ are not naturally in the same space, we align them using Procrustes alignment.
We hypothesize that using the embeddings from a period with an overall more positive (or negative) sentiment biases the classifier accordingly.
Thus, we estimate the sentiment shift between periods $t_i$ and $t_j$ with $\bar{d}_{i\leftarrow j} = \bar{s}_{i\leftarrow j} - \bar{s}_{i\leftarrow i}$. We conduct the experiment on two Croatian news sentiment analysis datasets: STONE \cite{baric-etal-2023-target}, comprising solely of news headlines, and 24sata \citep{pelicon_2020}, which focuses on full news articles.

To further validate the quality of word embeddings for sentiment drift, we also analyze the distribution of sentiment scores of news articles in each period.
Specifically, we sample 25k unlabeled articles per period from the TakeLab Retriever corpus.
To automatically assign sentiment labels, we train a transformer-based classifier using BERTić \cite{ljubesic-lauc-2021-bertic}, on the STONE and 24sata datasets, respectively. Further details on the training procedure and hyperparameter settings can be found in \Cref{subsec:sent_shift}.

\section{Results}

\subsection{Embedding Quality}
% \martin{Embedding quality: wordsim \& other word embedding quality measures, contrast to scores of other works (check papers citing those resources; copy results and contrast ours).  Goal is to show that the embeddings have sufficient quanity}

% ADD:
% - spearmann corr results (wordsim with crosemrel450)
% - Contrastive spread: synoynm vs non-related terms similarity diff using crosyn

We report the results of embedding quality evaluation in \Cref{tab:emb_eval}. % We assess quality of word embeddings using two different datasets. 
We measure the Spearman correlation between embedding-based cosine similarity and human judgments on the word similarity dataset CroSemRel450.  Additionally, we compute contrastive spread on the CroSYN dataset to evaluate how clearly word embeddings distinguish synonyms from unrelated words.
Focusing on nouns, adjectives, and verbs, we calculate the contrastive spread as the difference between a word's cosine similarity to its synonym and its similarity to an unrelated word, where higher scores reflect stronger semantic discrimination.
Overall, we find a moderate positive correlation of our estimated similarity with human judgments for word similarity across all periods. 
Both measurements indicate that embedding quality improves in later periods, highlighting the influence of data quantity on embedding quality. 
In contrast to similar embedding approaches for word similarity evaluation, our results are slightly worse albeit comparable ($\rho = 0.62$; \citet{zuanovic2014experiments}).
% In contrast to similar embedding approaches for word similarity evaluation, we obtain a comparable Spearman coefficient for Croatian embeddings ($\rho$ = 0.52; \citet{svoboda-beliga-2018-evaluation}).
%, but a lower correlation for English embeddings ($\rho$ = 0.71; \citet{wang2019embeddings}).}

\begin{table}[ht]
% \vspace{-0.5em}
\centering
\resizebox{\columnwidth}{!}{
\begin{tabular}{@{}c@{\hspace{3pt}}c@{\hspace{4pt}}c@{\hspace{4pt}}c@{\hspace{4pt}}c@{}}
\toprule
\multirow{2}{*}{\textbf{Period}} & \multirow{2}{*}{\textbf{Similarity ($\uparrow$})} & \multicolumn{3}{c}{\textbf{Contrastive spread} ($\uparrow$)} \\ \cmidrule(l){3-5} 
                &                  & \textbf{Noun} & \textbf{Adjective} & \textbf{Verb} \\ 
\midrule
1 (2000–2004)   & $0.49^{\dagger}$ & $0.08$ & $0.07$      & $0.05$ \\
2 (2005–2009)   & $0.49^{\dagger}$ & $0.14$ & $0.10$      & $0.09$ \\
3 (2010–2014)   & $0.52^{\dagger}$ & $0.21$ & $0.18$      & $0.15$ \\
4 (2015–2019)   & $0.51^{\dagger}$ & $0.26$ & $0.23$      & $0.19$ \\
5 (2020–2024)   & $0.51^{\dagger}$ & $0.27$ & $0.25$      & $0.21$ \\
\midrule
All (2000–2024) & $0.52^{\dagger}$ & $0.32$ & $0.27$      & $0.23$ \\
\bottomrule
\end{tabular}
}
\caption{Intrinsic embedding evaluation: word similarity ($^{\dagger}$\,=\,\textit{p}\,$<$\,0.001, Spearman correlation) and contrastive spread by period and part of speech.}
\vspace{-1em}
\raggedright
\label{tab:emb_eval}
\end{table}

% Result interpretation:
% - strong positive correlation for wordsim
% - Spearman correlation outperformed this (https://aclanthology.org/L18-1240.pdf) (only for wordsim) ???
% - contrastive spread: nouns showcase the highest contrast, meaning that they capture semantics better than verbs and adjectives, respectively ->maybe this could be the reason we focus on nouns in the subsequent analysis
% also, spread improves over time - more tokens, better quality

\subsection{Topical Linguistic Shift}
% \martin{Topical semantic shift: list out the major events in the region; categorize words to events (chatgpt; top5000?  10000?  words; translate (optional) \& assign to topics. Measure per-category shift.  Goal: how much did each event affect the narrative/discourse}

We provide a summary of words exhibiting most prominent shifts in \Cref{tab:topical_shift_summary}.
We show that neighboring words of top-shifting words inside a topic can pinpoint the period when words acquire new meanings.
We provide complete results of the top-picked shifting words inside each topic: \textsc{covid-19}, EU, and technology in \Cref{tab:topical_shift} in \Cref{subsec:tls_res}.
% Table \ref{tab:topical_shift} in Appendix \ref{subsec:tls_res} gives the complete results for the top-picked shifting words inside each topic: \textsc{covid-19}, EU, and technology.
% We show how neighboring words for top-shifting words inside a topic can pinpoint the period when the words acquired new meanings.
% Table \ref{tab:topical_shift_summary} summarizes the most prominent exemplars of the shift.

\begin{table*}
\centering
\adjustbox{width=\linewidth}{
\small
\begin{tabular}{ll p{2.5cm} p{2.5cm} p{2.5cm} p{2.5cm} p{2.5cm}}
\toprule

\multicolumn{2}{l}{\multirow{2}{*}{}} & \multicolumn{5}{c}{Top five noun neighbors} \\

\cmidrule(lr){3-7}

\multicolumn{1}{l}{Topic} & \multicolumn{1}{l}{Top shift} & {\textbf{2000--2004}} & \multicolumn{1}{c}{\textbf{2005--2009}} & \multicolumn{1}{c}{\textbf{2010--2014}} & \multicolumn{1}{c}{\textbf{2015--2019}} & \multicolumn{1}{c}{\textbf{2020--2024}} \\

\midrule

\multirow{-2}{*}{\rotatebox[origin=c]{90}{\shortstack{\scriptsize\textsc{Covid-19}}}}
& \parbox[c]{3cm}{varijanta \\ (variant) \\$D_c=0.53$}
& \parbox[c]{3cm}{\tiny
opcija (option)\\
kalkulator (calculator)\\
mogućnost (possibility)\\
solucija (solution)\\
opipavanje (palpation)
}
& \parbox[c]{3cm}{\tiny
opcija (option)\\
solucija (solution)\\
alternativa (alternative)\\
mogućnost (possibility)\\
verzija (version)
}
& \parbox[c]{3cm}{\tiny
opcija (option)\\
solucija (solution)\\
verzija (version)\\
inačica (version)\\
alternativa (alternative)
}
& \parbox[c]{3cm}{\tiny
verzija (version)\\
opcija (option)\\
solucija (solution)\\
alternativa (alternative)\\
vrsta (type, kind)
}
& \parbox[c]{3cm}{\tiny
delta (delta)\\
sojevi (strains)\\
mutacija (mutation)\\
podvrsta (subtype)\\
virus (virus)
} \\

\midrule

\multirow{-1}{*}{\rotatebox[origin=c]{90}{\shortstack{\scriptsize{EU}}}}  
& \parbox[c]{3cm}{fond \\ (fund) \\$D_c=0.32$}
& \parbox[c]{3cm}{\tiny
portfelj (portfolio)\\
kotacija (quotation)\\
benefit (benefit)\\
refinanciranje (refinancing)\\
socijala (poverty)
}
& \parbox[c]{3cm}{\tiny
alokacija (allocation)\\
benefit (benefit)\\
tranša (tranche)\\
uplaćivanje (payment) \\
stipendiranje (scholarship) 
}
& \parbox[c]{3cm}{\tiny
alokacija (allocation)\\
namicanje (raising)\\
kapital (capital)\\
dividenda (dividend)\\
banka (bank)
}
& \parbox[c]{3cm}{\tiny
financiranje (financing)\\
sufinanciranje (co-financing)\\
alokacija (allocation)\\
novac (money)\\
proračun (budget)
}
& \parbox[c]{3cm}{\tiny
alokacija (allocation)\\
ulaganje (investment)\\
sufinanciranje (co-financing)\\
obnova (renewal)\\
samofinanciranje (self-financing)
} \\

\midrule

\multirow{-1}{*}{\rotatebox[origin=t]{90}{\shortstack{\scriptsize{Tech}}}} 
& \parbox[c]{3cm}{inteligencija \\ (intelligence) \\$D_c=0.51$}
& \parbox[c]{3cm}{\tiny
kvocijent (quotient)\\
razlikovanje (distinction)\\
instinkt (instinct) \\
jasnoća (clarity)\\
evolucija (evolution)
}
& \parbox[c]{3cm}{\tiny
kvocijent (quotient)\\
empatija (empathy)\\
nadarenost (giftedness)\\
opažanje (perception)\\
habitus (habitus)
}
& \parbox[c]{3cm}{\tiny
kvocijent (quotient)\\
sposobnost (ability)\\
upućenost (familiarity)\\
racionalnost (rationality)\\
erudicija (erudition)
}
& \parbox[c]{3cm}{\tiny
algoritam (algorithm)\\
tehnologija (technology)\\
automatizacija (automation)\\
kvocijent (quotient)\\
robotika (robotics)
}
& \parbox[c]{3cm}{\tiny
tehnologija (technology)\\
algoritam (algorithm)\\
automatizacija (automation)\\
učenje (learning)\\
robotika (robotics)
} \\

\bottomrule
\end{tabular}}
\caption{Topical linguistic shift with respect to three topics: \textsc{Covid-19}, \textit{European Union (EU)}, and \textit{Technology (Tech)}. We pick one top shift noun word per topic based on the cumulative shift score (second column). For each of the picked words, we show the top five nearest noun neighbors over five periods. Translations are in parentheses.}
\label{tab:topical_shift_summary}
\vspace{-1em}
\end{table*}

\paragraph{\textsc{Covid-19}.} The \textsc{covid-19} crisis, which began in 2020, is reflected in the semantic shifts of words that were previously topically neutral, 
%As coronavirus struck the world during the fifth period, Croatia also felt its impact.
%This is reflected in the change of the neighboring words for seemingly topically neutral words,
such as \textit{maska (mask)} and \textit{varijanta (variant)}.
The word \textit{maska} changes from referring to a clothing item to an instrument for reducing viral transmission.
The noun \textit{varijanta} changes its dominant meaning during the fifth wave from an option or possibility to characterizing different strains (variants) of the coronavirus. % (alpha, beta, gamma, delta)
The word \textit{pandemija (pandemic)} changed a lot during the $25$ year period due to its connection to diverse diseases (from Ebola to flu and finally \textsc{Covid-19}). However, it was always used in the context of infectious diseases.

\paragraph{EU.} The evolution of EU-related terminology mirrors Croatia's path through three periods: considering EU membership, preparing for admission, and utilizing the benefits of being a member state.
The word \textit{integracija (integration)} changes from emphasizing bureaucratic \textit{harmonization} (2000--2004) to entering the \textit{union} (2013) and practical implementation and \textit{Europeanization} by 2020--2024.
\textit{Komisija (commission)} increasingly associates with legislative bodies such as the \textit{council}, \textit{ombudsman}, and \textit{parliament}, reflecting the importance of legal procedures for Croatia's admission into the EU.
Finally, \textit{fond (fund)} shifts from associating with financial terms such as \textit{quotation} and \textit{portfolio} to \textit{sufinanciranje (co-financing)} and \textit{obnova (renewal)} in the last two periods, reflecting usage of EU funds. %  for building a stronger future for Croatia

\paragraph{Technology.} Technological advancements are also reflected in linguistic shifts. \textit{Vjerodajnica (credential)} evolves from diplomatic words (\textit{delegation}, \textit{telegram}) to digital identifiers (\textit{password}, \textit{document}), signalling the transition into the digital era.
\textit{Inteligencija (intelligence)} changes from abstract cognitive attributes (\textit{quotient}, \textit{erudition}) to AI concepts (\textit{algorithms}, \textit{automation}), reflecting the post-2010 AI revolution.
Finally, \textit{privola (consent)} shifts from legal, in-person authorization to digital mechanisms such as \textit{kolačić (cookie)} and \textit{pohrana (data storage)}.

%\martin{Impact on sentiment: train sentiment classifier on pauza?  or some other Cro sentiment dataset (word2vec avg + logreg or word2vec + LSTM). Train classifiers using word vectors from each time period.  Then substitute for (aligned) vectors from another time period and see how much average sentiment changes (increase/decrease).  Goal: did overall sentiment change between time periods?  Also can be done by measuring frequency of words from positive/negative lexicons in news articles}

\subsection{Sentiment Shift}

\begin{figure}[ht]
    \centering
    \includegraphics[width=\linewidth]{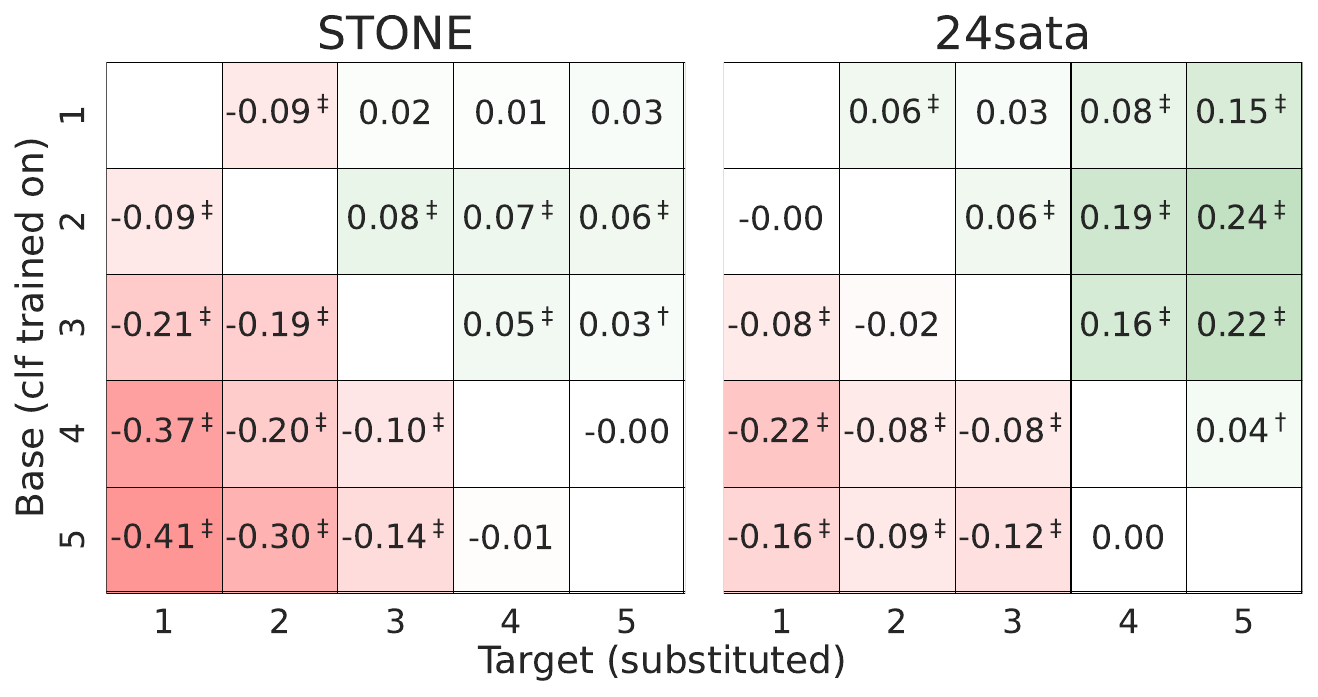}
    % \vspace{-2em}
    \caption{Sentiment shift between periods. Each cell ($i$,$j$) contains the value $\bar{d}_{i\leftarrow j}$.
    We compute statistical significance levels of the quantities being greater than zero using $10$-fold cross validation.  We denote $p < 0.05$ with $\dagger$ and  $p<0.01$ with $\ddagger$.}
    % \vspace{-1em}
    \label{fig:sentiment-both}
\end{figure}

We report results of sentiment shift on STONE and 24sata datasets in \Cref{fig:sentiment-both}.  We observe that transferring aligned embeddings from later periods into earlier periods increases average predicted sentiment, while the opposite holds when transferring embeddings from earlier periods to later. Additionally, we observe a similar trend regarding the increased share of positive words in more recent periods using a SentiLex lexicon for Croatian \cite{glavas-etal-2012-experiments}.

\begin{figure}[ht]
    \centering
    \includegraphics[width=\linewidth]{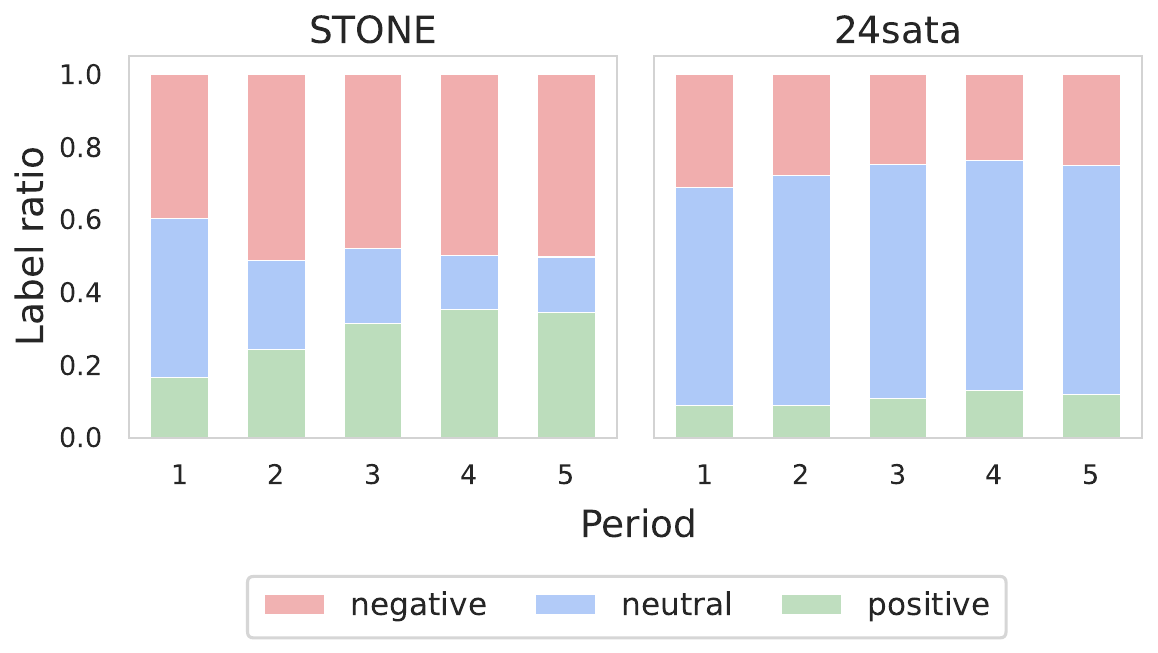}
    % \vspace{-2em}
    \caption{Change of predicted sentiment ratios when using classifiers trained on STONE and 24sata to categorize a sample of articles from Retriever. The trend of increased news polarization is more evident when using classifiers trained on STONE, but the same is evident for 24sata.}
    % \vspace{-1em}
    \label{fig:sentiment-check}
\end{figure}

We further investigate the increase in news positivity, through the distribution of sentiment labels for both news headlines and full articles across different time periods in \Cref{fig:sentiment-check}.
We find that in general, the amount of articles labeled as positive increases at the expense of neutral ones. The proportion of negative labels also slightly increased over time, particularly in news headlines. 
These results corroborate the findings of sentiment shift, indicating an increase of positivity in news in recent periods. 

We hypothesize that increased positivity in news may be driven by one of several phenomena observed in media communication.
Increased positivity could be the a reaction to general negativity, influenced by the decline of mental health in the general population \citep{rozanov2019mental,cullen2020mental}.
The increase in positivity could also be attributed to online news covering more diverse, less serious topics, or the increase in satirical or comedic articles.
Another potential factor is the increased polarization of media discourse, where news content is becoming more extreme in its use of emotionally charged language to elicit reactions from readers \cite{rozado2022news}.
Nonetheless, we believe that this phenomenon, in which sentiment expressed in news articles contrasts broader negativity, warrants further study as it may affect the quality of models trained on corpora from different time periods.

\section{Conclusion}

We apply diachronic word embedding analysis to Croatian, a language scarce in historical corpora.
By training diachronic embeddings on Croatian online news articles spanning the last $25$ years, we successfully detect linguistic shifts pertaining to recent major events, exhibited by existing words acquiring new meanings or completely changing how they are used. These results show that linguistic shifts can also be detected in shorter time spans.
We also reveal practical implications of linguistic shifts on sentiment analysis, showing that 
%  sentiment analysis of trained embeddings reveals that 
word meanings from recent periods tend to be more positive, contrasting with research indicating an increase in negativity over the same period.
%\newpage 

%While previous research has used diachronic word embeddings to identify shifts in word meaning for higher-resource languages such as English, we apply the same approach to Croatian, a language with fewer available resources.

\section*{Limitations}

In our experiments, we analyze only five-year periods, revealing some regularities that might be too coarse- or fine-grained for others. We experimented with two-year periods but found them too fine-grained. Future works can vary the duration of periods.
% Further insights may be gained from more fine-grained splits, such as two-year periods.
We use a lexicon-based context-free lemmatizer (MOLEX), which could be error-prone and introduce noise to the experiments. 
The distribution of article count per period varies significantly as earlier periods have fewer articles. This fact influences the quality of produced word embeddings and could bias the results.
Finally, we only explore a single distributed word embedding method in SGNS, the results of which need not generalize to other methods.

%Next, we use MOLEX for lemmatization, which returns only one lemma for each word, irrespective of the context, which could have introduced errors.
% Silver-labeled part-of-speech tags could also have affected the quality of the individual words for training.
% While part-of-speech tagging aids in distinguishing homonyms, it does not explicitly differentiate polysemous words, which share a single grammatical category but carry multiple related meanings. % \todo{add other limitations}

% \section*{Acknowledgments}

\bibliography{anthology,custom}
% \bibliography{custom}

\appendix

\section{Topical Linguistic Shift} 
\label{subsec:tls_res}

\paragraph{Analyzing Topical Linguistic Shift Embeddings.}

We create a common vocabulary between periods to measure only the words with sufficient frequency in each period.
In total, the five periods share \num{348679} words.
We curate a list of potential word shifters for each topic (both top shifters by $D_\mathrm{c}$ and additional words we expected to shift).
We remove words with frequency less than \num{1000} over $25$ years for each topic list separately.
Next, for each word from the curated topic list, we compute $D_c$ and pick $20$ candidates with the highest shift score for further analysis.
For each candidate, we find its nearest \num{1000} noun neighbors by cosine distance.
Out of these \num{1000}, we pick $20$ that occur at least $20$ times in each period. We analyze the linguistic shift of words using their neighbors for each topic and pick the most interesting words representing the topic shift with its closest and most representative neighbors.

\paragraph{Full Topical Shift Results.}
We report full results of topical shift on terms pertaining to major events in \Cref{tab:topical_shift}.

\section{Sentiment Shift} 
\label{subsec:sent_shift}

\paragraph{Training Setup for Sentiment Classifiers}

To train the sentiment classifier for Croatian news, we use the STONE and 24sata datasets with the BERTić model \cite{ljubesic-lauc-2021-bertic}.
For the STONE dataset, we utilize only the tone labels, as they capture the overall tone of the headline, aligning with our definition of sentiment. We achieve an F1 score of $0.77$ on STONE and $0.73$ on the 24sata dataset.

\section{Hyperparameters and Hardware Details}
\label{app:hyperparams}
We train word embeddings and the sentiment regressor on a machine with 2x AMD Epyc 7763 CPUs and 512 GB of RAM. In \autoref{tab:hyperparams}, we report the hyperparameters used in word embedding training. Our setup mostly follows that of \citet{hamilton-etal-2016-diachronic} with a key difference that we do not restrict our vocabulary but train the embeddings on all the words in the corpus. We discard only punctuation words identified by a part-of-speech tagger and lowercase all the words before training. When training classifiers for sentiment analysis (cf.~\sect{sec:senti-shift}), we use the implementation of logistic regression from scikit-learn \cite{pedregosa2011scikit} with the default hyperparameters.

We train both BERTić sentiment classifiers on an NVIDIA RTX 3090 GPU with 24GB RAM using CUDA 12.9 and the HuggingFace Trainer\footnote{\url{https://huggingface.co/docs/transformers/main_classes/trainer}} library. We employ the default hyperparameters provided by the Trainer and train for $3$ epochs with a batch size of $8$.

\begin{table}[]
    \centering
    \addtolength{\tabcolsep}{22pt}    
    \begin{tabular}{cc}
    \toprule
     \textbf{Hyperparameter}    & \textbf{Value} \\
     \midrule
         \texttt{vector\_size} & \texttt{300} \\
         \texttt{window} & \texttt{4} \\
         \texttt{negative} & \texttt{5} \\
         \texttt{sample} & \texttt{1e-5} \\
         \texttt{alpha} & \texttt{0.02} \\
         \texttt{epochs} & \texttt{5} \\
    \bottomrule
    \end{tabular}
    \caption{Hyperparameters for word embedding training. The names of hyperparameters in the first column match the argument names when initializing a \textsc{Gensim} Word2Vec model.}
    \label{tab:hyperparams}
\end{table}

\begin{table*}
\centering
\adjustbox{width=1.0\linewidth}{
% \rotatebox[origin=c]{270}{
\small
\begin{tabular}{l p{1.5cm} p{2.5cm} p{2.5cm} p{2.5cm} p{2.5cm} p{2.5cm}}
\toprule

\multicolumn{2}{l}{\multirow{2}{*}{}} & \multicolumn{5}{c}{Top five noun neighbors} \\

\cmidrule(lr){3-7}

\multicolumn{1}{l}{Topic} & \multicolumn{1}{l}{Top shift words} & {\textbf{2000--2004}} & \multicolumn{1}{c}{\textbf{2005--2009}} & \multicolumn{1}{c}{\textbf{2010--2014}} & \multicolumn{1}{c}{\textbf{2015--2019}} & \multicolumn{1}{c}{\textbf{2020--2024}} \\

\midrule

\multirow{9}{*}{\rotatebox[origin=c]{90}{\shortstack{\textsc{Covid-19}}}}  
% maska_noun row
& \parbox[c]{3cm}{maska \\ (mask) \\ $D_c=0.66$}
& \parbox[c]{3cm}{\tiny
kabanica (raincoat)\\
kombinezon (coverall)\\
značka (badge)\\
lampica (little lamp)\\
šminka (makeup)
}
& \parbox[c]{3cm}{\tiny
lice (face)\\
šilterica (visor cap)\\
kombinezon (coverall)\\
šešir (hat)\\
frak (tailcoat)
}
& \parbox[c]{3cm}{\tiny
rukavica (glove)\\
šminka (makeup)\\
šešir (hat)\\
pancirka (flak jacket)\\
štitnik (protector, shield)
}
& \parbox[c]{3cm}{\tiny
perika (wig)\\
povez (band, patch)\\
šminka (makeup)\\
kaciga (helmet)\\
štitnik (protector, shield)
}
& \parbox[c]{3cm}{\tiny
nošenje (wearing)\\
rukavica (glove)\\
nenošenje (not wearing)\\
distanca (distance)\\
pleksiglas (plexiglass)
} \\

\cmidrule(lr){2-7}

& \parbox[c]{3cm}{pandemija \\ (pandemic) \\ $D_c=0.61$}
& \parbox[c]{3cm}{\tiny
ebola (Ebola)\\
incidencija (incidence)\\
ospice (measles)\\
virus (virus)\\
epidemiolozi (epidemiologists)
}
& \parbox[c]{3cm}{\tiny
SARS (SARS)\\
ebola (Ebola)\\
gripa (flu)\\
ospice (measles)\\
pojavnost (prevalence)
}
& \parbox[c]{3cm}{\tiny
SARS (SARS)\\
ebola (Ebola)\\
ospice (measles)\\
kuga (plague)\\
virolog (virologist)
} 
& \parbox[c]{3cm}{\tiny
ebola (Ebola)\\
bolest (disease)\\
ospice (measles)\\
kriza (crisis)\\
epidemiolozi (epidemiologists)
}
& \parbox[c]{3cm}{\tiny
epidemija (epidemic)\\
korona (corona)\\
kriza (crisis)\\
lockdown (lockdown)\\
\textsc{covid} (\textsc{covid})
} \\

\cmidrule(lr){2-7}

& \parbox[c]{3cm}{varijanta \\ (variant) \\ $D_c=0.53$}
& \parbox[c]{3cm}{\tiny
opcija (option)\\
kalkulator (calculator)\\
mogućnost (possibility)\\
solucija (solution)\\
opipavanje (palpation)
}
& \parbox[c]{3cm}{\tiny
opcija (option)\\
solucija (solution)\\
alternativa (alternative)\\
mogućnost (possibility)\\
verzija (version)
}
& \parbox[c]{3cm}{\tiny
opcija (option)\\
solucija (solution)\\
verzija (version)\\
inačica (version)\\
alternativa (alternative)
}
& \parbox[c]{3cm}{\tiny
verzija (version)\\
opcija (option)\\
solucija (solution)\\
alternativa (alternative)\\
vrsta (type, kind)
}
& \parbox[c]{3cm}{\tiny
delta (delta)\\
sojevi (strains)\\
mutacija (mutation)\\
podvrsta (subtype)\\
virus (virus)
} \\

\midrule

\multirow{9}{*}{\rotatebox[origin=c]{90}{\shortstack{European Union}}}  

& \parbox[c]{3cm}{integracija \\ (integration) \\ $D_c=0.39$}
& \parbox[c]{3cm}{\tiny
harmonizacija (harmonization)\\
agenda (agenda) \\ 
aspirant (aspirant) \\
iskorak (step forward) \\
kohezija (cohesion)
}
& \parbox[c]{3cm}{\tiny
unija (union)\\
agenda (agenda)\\
fragmentacija (fragmentation)\\
dobrosusjedstvo (neighborliness) \\
ulazak (entrance)
}
& \parbox[c]{3cm}{\tiny
unija (union)\\
implementacija (implementation)\\
razvoj (development)\\
dobrosusjedstvo (neighborliness)\\
međuovisnost (interdependence)
}
& \parbox[c]{3cm}{\tiny
unija (union)\\
razvoj (development)\\
inkluzija (inclusion)\\
povezivanje (connection) \\
jačanje (strengthening)
}
& \parbox[c]{3cm}{\tiny
implementacija (implementation)\\
uključenost (inclusion)\\
povezivanje (connection)\\
razvoj (development)\\
europeizacija (Europeanization)
} \\

\cmidrule(lr){2-7}

& \parbox[c]{3cm}{komisija \\ (commission) \\ $D_c=0.34$}
& \parbox[c]{3cm}{\tiny
ombudsman (ombudsman)\\
sukladnost (compliance)\\
delegacija (delegation)\\
unija (union)\\
nacrt (draft)
}
& \parbox[c]{3cm}{\tiny
delegacija (delegation)\\
unija (union)\\
mjerodavnost (competence)\\
arbitraža (arbitration)\\
instancija (instance)
}
& \parbox[c]{3cm}{\tiny
unija (union)\\
monitoring (monitoring)\\
ombudsman (ombudsman)\\
povjerenik (commissioner) \\
vijeće (council)
}
& \parbox[c]{3cm}{\tiny
unija (union)\\
prijedlog (proposal)\\
vlada (government)\\
parlament (parliament)\\
vijeće (council)
}
& \parbox[c]{3cm}{\tiny
unija (union)\\
smjernica (guideline)\\
vlada (government)\\
članica (member)\\
parlament (parliament)
} \\

\cmidrule(lr){2-7}

& \parbox[c]{3cm}{fond \\ (fund) \\ $D_c=0.32$}
& \parbox[c]{3cm}{\tiny
portfelj (portfolio)\\
kotacija (quotation)\\
benefit (benefit)\\
refinanciranje (refinancing)\\
socijala (poverty)
}
& \parbox[c]{3cm}{\tiny
alokacija (allocation)\\
benefit (benefit)\\
tranša (tranche)\\
uplaćivanje (payment) \\
stipendiranje (scholarship) 
}
& \parbox[c]{3cm}{\tiny
alokacija (allocation)\\
namicanje (raising)\\
kapital (capital)\\
dividenda (dividend)\\
banka (bank)
}
& \parbox[c]{3cm}{\tiny
financiranje (financing)\\
sufinanciranje (co-financing)\\
alokacija (allocation)\\
novac (money)\\
proračun (budget)
}
& \parbox[c]{3cm}{\tiny
alokacija (allocation)\\
ulaganje (investment)\\
sufinanciranje (co-financing)\\
obnova (renewal)\\
samofinanciranje (self-financing)
} \\

\midrule

\multirow{9}{*}{\rotatebox[origin=c]{90}{\shortstack{Technology}}}  
& \parbox[c]{3cm}{vjerodajnica \\ (credential) \\ $D_c=0.56$}
& \parbox[c]{3cm}{\tiny
otpravnik (ambassador's deputy)\\
delegacija (delegation)\\
diplomat (diplomat)\\
brzojav (telegram) \\
monsinjor (monsignor)
}
& \parbox[c]{3cm}{\tiny
otpravnik (ambassador's deputy)\\
diplomat (diplomat)\\
useljeništvo (immigration)\\
podtajnik (undersecretary)\\
parafiranje (initialing)
}
& \parbox[c]{3cm}{\tiny
telefaks (fax)\\
adresar (address book)\\
ovjera (certification)\\
fotokopija (photocopy)\\
tiskanica (form)
}
& \parbox[c]{3cm}{\tiny
formular (form)\\
iskaznica (ID card)\\
brzojav (telegram)\\
pošta (mail)\\
veleposlanik (ambassador)
}
& \parbox[c]{3cm}{\tiny
građani (citizens)\\
iskaznica (ID card)\\
lozinka (password)\\
putovnica (passport)\\
dokument (document)
} \\

\cmidrule(lr){2-7}

& \parbox[c]{3cm}{inteligencija \\ (intelligence) \\ $D_c=0.51$}
& \parbox[c]{3cm}{\tiny
kvocijent (quotient)\\
razlikovanje (distinction)\\
instinkt (instinct) \\
jasnoća (clarity)\\
evolucija (evolution)
}
& \parbox[c]{3cm}{\tiny
kvocijent (quotient)\\
empatija (empathy)\\
nadarenost (giftedness)\\
opažanje (perception)\\
habitus (habitus)
}
& \parbox[c]{3cm}{\tiny
kvocijent (quotient)\\
sposobnost (ability)\\
upućenost (familiarity)\\
racionalnost (rationality)\\
erudicija (erudition)
}
& \parbox[c]{3cm}{\tiny
algoritam (algorithm)\\
tehnologija (technology)\\
automatizacija (automation)\\
kvocijent (quotient)\\
robotika (robotics)
}
& \parbox[c]{3cm}{\tiny
tehnologija (technology)\\
algoritam (algorithm)\\
automatizacija (automation)\\
učenje (learning)\\
robotika (robotics)
} \\

\cmidrule(lr){2-7}

& \parbox[c]{3cm}{privola \\ (consent) \\ $D_c=0.50$}
& \parbox[c]{3cm}{\tiny
staratelj (guardian)\\
očevidnik (register)\\
autorizacija (authorization)\\
ovlaštenje (authorization) \\
pozivatelj (caller)
}
& \parbox[c]{3cm}{\tiny
uvjetovanje (conditioning)\\
obvezivanje (commitment)\\
direktiva (directive)\\
konzultiranje (consultation) \\
suodlučivanje (co-decision)
}
& \parbox[c]{3cm}{\tiny
pohrana (storage)\\
odobrenje (approval)\\
ustanoviti (establish)\\
suglasnost (accord) \\
supotpis (co-signature)
}
& \parbox[c]{3cm}{\tiny
pohrana (storage)\\
kolačić (cookie)\\
povjerljivost (confidentiality)\\
suglasnost (accord) \\
stranica (page)
}
& \parbox[c]{3cm}{\tiny
pohrana (storage)\\
suglasnost (accord)\\
kolačić (cookie)\\
dopuštenje (permission)\\
stranica (page)
} \\

\bottomrule
\end{tabular}
}
% }
\caption{Full topical linguistic shift results with respect to three topics: \textsc{Covid-19}, \textit{European Union}, and \textit{Technology}. We pick three top shift noun words per topic based on the cumulative shift score (second column). For each of the picked words, we show the top five nearest noun neighbors over five periods. Translations are in parentheses.}
\label{tab:topical_shift}
\end{table*}

\end{document}